\documentclass[runningheads]{llncs}
\usepackage[T1]{fontenc}
\usepackage{graphicx}
\usepackage[listings]{tcolorbox}
\tcbuselibrary{listings,theorems}
\definecolor{purple}{HTML}{8d3a94}
\definecolor{c3}{HTML}{fe793d}
\definecolor{red1}{HTML}{FFCCC9}
\definecolor{green1}{HTML}{B5FFAA}

\usepackage{multirow}

\newtcbtheorem[]{exmp}{Prompt}%
{colback=purple!5,colframe=purple!80,fonttitle=\bfseries, left=.02in, right=.02in,bottom=.02in, top=.02in}{exmp}

\newtcbtheorem[]{exmp2}{Hint}%
{colback=c3!5,colframe=c3!80,fonttitle=\bfseries, left=.02in, right=.02in,bottom=.02in, top=.02in}{exmp2}

\usepackage{xcolor}
\usepackage{colortbl}
\usepackage{color}
\usepackage{graphicx}
\usepackage{enumerate}
\usepackage{subcaption}
\usepackage{float}

\usepackage{booktabs}

\usepackage{algorithm}
\usepackage{algorithmic}

\begin{document}
\title{TUMS: Enhancing Tool-use Abilities of LLMs with Multi-structure Handlers}
%
%
\author{Aiyao He\inst{1,2,3}\orcidID{0009-0002-7982-4442} \and
Sijia Cui\inst{1,4}\orcidID{0009-0004-7304-9064} \and
Shuai Xu\inst{2,3,5}\orcidID{0009-0009-0915-3891} \and
Yanna Wang \inst{1}\orcidID{0000-0002-4135-8442} \and
Bo Xu\inst{1,2,*}\orcidID{0000-0002-8884-0447}}
\authorrunning{A. He et al.}
%
\institute{
Institute of Automation, Chinese Academy of Sciences, Beijing, China \\
\email{\{heaiyao2022, cuisijia2022, wangyanna2013, boxu\}@ia.ac.cn} \and
Nanjing Artificial Intelligence Research of IA, Nanjing, China \and
University of Chinese Academy of Sciences,Nanjing, Nanjing, China \\ 
\email{xushuai23@mails.ucas.ac.cn} \and
School of Artificial Intelligence, University of Chinese Academy of Sciences, \\Beijing, China \and
Nanjing University of Information Science \& Technology, Nanjing, China \\
*Corresponding Author: Bo Xu, \email{boxu@ia.ac.cn}
}
\maketitle              
\begin{abstract}
Recently, large language models(LLMs) have played an increasingly important role in solving a wide range of NLP tasks, leveraging their capabilities of natural language understanding and generating. Integration with external tools further enhances LLMs' effectiveness, providing more precise, timely, and specialized responses. 
However, LLMs still encounter difficulties with non-executable actions and improper actions, which are primarily attributed to incorrect parameters. The process of generating parameters by LLMs is confined to the tool level, employing the coarse-grained strategy without considering the different difficulties of various tools.
To address this issue, we propose TUMS, a novel framework designed to enhance the tool-use capabilities of LLMs by transforming tool-level processing into parameter-level processing. Specifically, our framework consists of four key components: (1) an intent recognizer that identifies the user's intent to help LLMs better understand the task; (2) a task decomposer that breaks down complex tasks into simpler subtasks, each involving a tool call; (3) a subtask processor equipped with multi-structure handlers to generate accurate parameters; and (4) an executor.
Our empirical studies have evidenced the effectiveness and efficiency of the TUMS framework with an average of 19.6\% and 50.6\% improvement separately on easy and hard benchmarks of ToolQA, meanwhile, we demonstrated the key contribution of each part with ablation experiments, offering more insights and stimulating future research on Tool-augmented LLMs.

\end{abstract}

\section{Introduction}

During pre-training, the large language models (LLMs), such as GPT series\cite{achiam2023gpt4}, Llama\cite{touvron2023llama}, and Qwen\cite{qwen1.5}, acquire extensive knowledge from a wide variety of corpora. Compared to smaller language models, LLMs exhibit surprising emergent abilities such as in-context learning\cite{dong2022icl_survey}, instruction following\cite{ouyang2022instructGPT}, and step-by-step reasoning\cite{wei2022cot}, enhancing their performance in complex and specific NLP tasks. 
However, LLMs have limitations, including inaccurate outputs on complex problems, lack of access to the latest information, and potential negative consequences due to memorizing parts of the training data\cite{carlini2023quantifying}. These issues cannot be resolved simply by increasing the scale of training. 

\begin{figure}[t]
\centering
\centerline{\includegraphics[width=\textwidth]{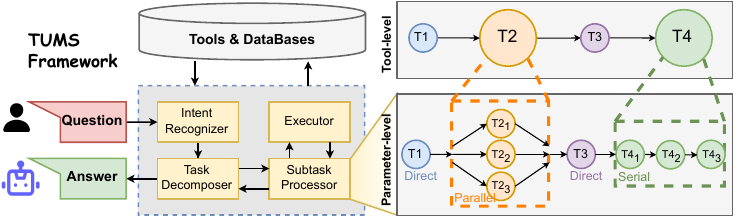}}
\caption{The illustration of \textbf{Tool-level} and \textbf{Parameter-level}. 
The classical tool-use workflow on QA tasks only reaches the depth of the tool level, while our method, TUMS, has a fine-grained treatment on the parameter level.}
\label{fig:tums}
\end{figure}

Using external tools serves as an effective approach to address these issues. Research has achieved remarkable results in various tool usages, such as retrieval tools\cite{gao2023retrieval_survey,zhao2024retrieval}, calculator tools\cite{cobbe2021gsm8k,parisi2022talm}, code tools\cite{chen2021codex,singh2023progprompt,yang2024code_survey}, query tools\cite{jiang2023structgpt}, multimodal tools\cite{gupta2023visual,suris2023vipergpt}, \textit{etc.} 
As task complexity increases, LLMs need to invoke multiple tools to complete tasks, which requires LLMs to select tools from the toolset and decide the execution sequence.
After tool selection and tool planning, LLMs generate corresponding parameters for these tool functions.
Despite many studies making progress in improving parameter accuracy\cite{qin2023tool_learning,hsieh2023tool_doc,liu2023controlllm,shi2024coagnet,hao2024toolkengpt}, LLMs still face serious issues with missing parameters, generating irrelevant parameters, and producing incorrect parameters. 
The cause of this issue is rooted in the limitation that the parameter generation procedure within LLMs is restricted to the tool level, which is characterized by a uniform generation structure. 
As depicted in Figure \ref{fig:tums}, the complexity of utilizing various tools differs. 
It is evident that a singular parameter generation structure is incapable of producing error-free parameters for various tools, thereby leading to non-executable actions and improper actions.

Similarly, humans do not adopt a one-size-fits-all strategy when confronting tasks of diverse complexities. To exemplify this, consider the hypothetical task of placing an elephant into a refrigerator, which can be conceptually segmented into three distinct phases: (1)open the refrigerator, (2) put in the elephant, (3) close the refrigerator. It is crucial to recognize that the degrees of difficulty associated with these steps are not equivalent. The actions of opening and closing the refrigerator are relatively straightforward and can be performed with directness. In contrast, the act of placing the elephant inside the refrigerator represents a more complex, coarse-grained step that defies direct execution. This step necessitates a further decomposition into a series of finer-grained subtasks, each of which must be meticulously addressed to achieve the overarching goal.

\textbf{Inspired by the patterns of human thought, we introduce TUMS, a novel framework for enhancing LLMs' tool-use abilities by employing parameter-level processing. }
Our framework consists of four key components: an intent recognizer, a task decomposer, a subtask processor, and an executor.
Figure \ref{fig:tums_full} illustrates the entire workflow of our framework in detail.
Initially, the intent recognizer identifies the user's intent by extracting keywords from the question and then generates a hint containing information about the related corpus and optional tools. The hint is passed to the task decomposer, helping LLMs better understand the task. 
The task decomposer then decomposes the complex question into simpler subtasks and assigns a corresponding tool.
After the task decomposer generates a tool-subtask pair, the subtask processor is responsible for generating parameters for the tool function and returning an answer by executing the tool in the executor.
In particular, we have taken into account the difficulty of tools and the number of parameters, dividing all tools into multiple types, each of which can be addressed by the corresponding structure handler. The details of our pipeline can be found in Section \ref{sec:3.3}.

Experimentally, we evaluate some existing methods and TUMS on ToolQA, and results show that TUMS has almost an overall enhancement at eight datasets with an average of 19.6\% improvement on easy questions and an average of 50.6\% on hard questions. TUMS-PRE, a TUMS variant utilizing preference-based hints shown in Appendix \ref{apd:second}, can achieve a 40.13\% correct rate across all questions, surpassing the 29.93\% correct rate of ReAct.

In summary, the main contributions are as follows:
\begin{enumerate}[(1)]
    \item We propose TUMS, a framework that consists of four components to enhance the tool-use capabilities of LLMs for complex question-answering tasks.
    \item We design multi-structure handlers for the subtask processor, transforming parameter generation from the tool level to the parameter level.
    \item The experiment results indicate the superiority of our framework, providing experimental evidence of the advantages of multi-structure, which inspires future further research. 
\end{enumerate}

\section{Related Work}

\subsection{Language models for reasoning}

In the standard prompting case, LLMs directly generate the answer to the question. Chain-of-Thought(CoT) prompting\cite{wei2022cot} proposes a method for enabling LLMs to generate a chain of thought by adding a series of intermediate reasoning steps, which significantly enhances the LLMs' ability to perform complex reasoning tasks. Inspired by CoT, CoT-SC\cite{wang2022self}, Tree of Thoughts(ToT)\cite{yao2024tree} and Graph of Thoughts(GoT)\cite{besta2024graph} enrich the reasoning processes of LLMs by respectively proposing multiple chains of thought, tree structure of thought, and graph structure of thought.

Question decomposition is another reasoning method that divides the original question into multiple subtasks, thereby effectively enhancing the reasoning capabilities of LLMs\cite{radhakrishnan2023QuestionDecomposition}. Least-to-most prompting\cite{zhou2022least}  first generates subtasks and then solves them sequentially, while successive prompting\cite{dua2022successive} and decomposed prompting\cite{khot2022decomposed} iteratively generate new subtasks based on previous answers.
These approaches have yielded substantial enhancements in a variety of symbolic reasoning and textual tasks. Drawing on the concept of task decomposition, our proposed framework, TUMS, decomposes complex tasks into several subtasks, each of which is interfaced with an external tool. This enables the resolution of issues that are more complex and have a wider range of applications.

\subsection{Tool-augmented language models}

Augmenting large language models with external tools can mitigate the issue of hallucination, provide access to up-to-date information, and enhance the accuracy of complex tasks. 
As task complexity grows, this approach has gradually evolved from single-tool invocation\cite{parisi2022talm} to multi-tool invocation, utilizing the reasoning ability of LLMs. 
HuggingGPT\cite{shen2024hugginggpt}, Chameleon\cite{lu2024chameleon}, ViperGPT\cite{suris2023vipergpt}, Visual ChatGPT\cite{wu2023visual} are methods that enable LLMs to sequentially produce all necessary tools in a single step. By contrast, ReAct\cite{yao2022react} and RestGPT\cite{song2023restgpt} adaptively generate the subsequent tool based on the results of previous tool executions, allowing for intermediate feedback after each action.
Recent studies have introduced tool retrieval\cite{kong2023tptu-v2,huang2024pluto,yuan2023craft,du2024anytool,qu2024colt}, tree\cite{qin2023toolllm,zhuang2023toolchain} or graph\cite{liu2023controlllm,liu2024toolnet} search to improve the efficiency and correctness of tool selection. Several works perform self-verification\cite{mekala2024toolverifier} or iterative optimization\cite{shi2024conAgents,chen2024smurfs} to improve parameter generation.
However, the processing depth of these methods is confined to the tool level, neglecting the variance in complexity and parameter amounts among different tools. Our approach refines the depth to the parameter level, selecting appropriate parameter generation structures for different tools, thereby enhancing the tool-use ability of LLMs.

\section{Method}

\begin{algorithm}[t]
\caption{The Inference Process of \textbf{TUMS}}
\label{alg:tums}
\begin{algorithmic}[1]
    \REQUIRE The question $Q$, Large Language Model $L$, prompt of recognizer $P^r$, prompt of decomposer $P^d$, prompt set of multi-structures handlers $\mathcal{P}^h$, $\mathcal{P}^h_f$ is a prompt of tool $f$ with specific structure, $\mathcal{H}_c$ is the hint about dataset $c$, parser function $g$.
    \ENSURE The Answer $A$
    \STATE \# Recognizer recognizes intent and generates hint.
    \STATE Recognizer: $c \sim L(\cdot|P^r,Q)$
    \STATE \# Decomposer collaborates with Processor and Executor.
    \FOR{$i=1,2,\dots,\text{MAX-STEPS}$}
        \STATE Decomposer: $s_{i} \sim L(\cdot|P^d,Q,\mathcal{H}_c,s_1,r_1,\dots,s_{i-1},r_{i-1})$
        \STATE $f_i,Q_i \sim g(s_i)$ \space \#  extract tool $f_i$ and subtask $Q_i$ from response of decomposer
        \STATE Processor: $a_i \sim L(\cdot|\mathcal{P}^h_{f_i},f_i,Q_i)$ \# generate accurate parameter $a_i$
        \IF{$f_i==\text{finish}$}
        \STATE $A=a_i$; \space Break
        \ENDIF
        \STATE Executor: $r_i = \text{tool-calling}(f_i,a_i)$ \# execute tool $f_i$ with parameter $a_i$
    \ENDFOR
\end{algorithmic}
\end{algorithm}

\begin{figure}[h]
\centering
\centerline{\includegraphics[width=\textwidth]{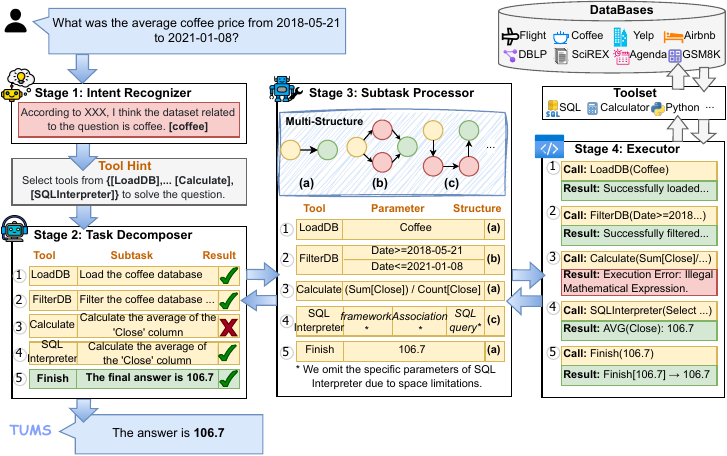}}
\caption{The proposed framework: TUMS.}
\label{fig:tums_full}
\end{figure}

LLMs exhibit strong abilities in in-context learning. 
Utilizing few-shot prompting techniques, we enable LLMs within our proposed framework, TUMS, to discern user intentions, decompose intricate tasks, and properly invoke tools to address each subtask effectively.
As depicted in Figure \ref{fig:tums_full}, TUMS comprises four modules: an intent recognizer, a task decomposer, a subtask processor, and an executor. Next, we will provide a detailed description of each module and prompts for all modules can be found in Figure \ref{fig:prompts_hints}. The overall inference process of the TUMS framework is detailed in Algorithm \ref{alg:tums}.

\subsection{Intent Recognizer} \label{sec:3.1}
Our framework begins with intent recognizer, a stage for discerning the user's underlying intent. This module can capture salient keywords and deduce the nature of the user's inquiry, encompassing a spectrum of query types such as database queries, graphical data analyses, information retrieval on text corpus, and mathematical problem-solving. By identifying the user's intent, we narrow down the toolset required for addressing the posed question, specifically, append a hint to the prompt for subsequent task decomposer, as shown in Figure \ref{fig:prompts_hints}.

\subsection{Task Decomposer} \label{sec:3.2}
After gaining a general understanding of the user's intent, the process enters the task decomposer, which is responsible for breaking down the original question into simpler subtasks and selecting an appropriate tool for each subtask. This task decomposition is an online process that requires iterative interaction with the subtask processor. Specifically, in each iteration, the decomposer generates one subtask and selects one corresponding tool. They are then handed over to the processor, which uses the specific tool to complete the subtask and then returns the result. Based on the previous collection of subtask-result trajectories, the decomposer determines whether further decomposition is needed. If the question has been settled\footnote{Once decomposer invoking the [Finish] tool.} or the iteration reaches the maximum steps, the decomposition terminates, with the final answer obtained by parsing from the last result. This decomposer module of TUMS is capable of making reflective adjustments to rectify the mistakes or select alternative tools, as shown in Figure \ref{fig:tums_full}.

\subsection{Subtask Processor}  \label{sec:3.3}
The essential difference between the proposed TUMS and other existing methods is the processor, in which TUMS applies multi-structure handlers with deliberation to generate the more accurate parameter of tool-calling. 


In each iteration, once the task decomposer has produced a subtask and selected a tool, the subtask processor needs to generate the requisite parameters for this tool to form a complete tool function, which is then conveyed to the executor to obtain the execution results.
As mentioned before, the accuracy of tool invocation is influenced by two principal factors: the number of parameters involved and the degree of difficulty in using the tools.
Focusing on improving the quality of parameters, we use a divide-and-conquer strategy that heuristically decomposes the parameter-generation process based on the subjective complexity of the tool.
We consider three scenarios of tool invocation: (1) invocation of simple tools with limited parameters; (2) invocation of simple tools with numerous parameters; and (3) invocation of intricate tools.
The others do not take into account these three scenarios, only decomposing to the tool level and adopting a uniform structure for parameter generation. 
Such a coarse-grained process is prone to errors, especially in scenarios with multi-parameter or intricate tools, which can lead to a cascade of subsequent errors.
We employ a fine-grained approach that further refines the tool level, utilizing a multi-structured parameter generation method to transform the tool level into the parameter level, thereby effectively addressing the aforementioned three scenarios of tool invocation.
Figure \ref{fig:tums_full} illustrates three structures for parameter generation.

\textbf{(a) Direct Generation Structure.} Given the subtask and the selected tool, the subtask processor generates a parameter directly. 

\textbf{(b) Parallel Generation Structure.} This structure is designed for situations with multiple parameters, based on the idea of splitting before combining. For example, in a data filtering task of the flight database, the processor first categorizes the filtering information in the question by time, space and object, which correspond respectively to flight dates, cities involved, and flight numbers. Following this, it identifies the corresponding filtering conditions for each category. Finally, integrate all conditions to form the ultimate parameters.

\textbf{(c) Serial Generation Structure.} Generating appropriate parameters for complex tool invocations, such as writing SQL queries, poses a challenge. However, this complexity is mitigated by breaking down the process into a series of manageable sequential steps.
Therefore, we can decompose the SQL query subtask further. First, extract the framework of the SQL query statement. Second, map the implicit information in the natural language queries to the explicit column names and their corresponding values in the database. Finally, synthesize the results of the previous two steps to generate the final SQL query statement.

The processor of TUMS is capable of expansion to more sophisticated structures, allowing for the tackling of a greater variety of complex tools.

\subsection{Executor} \label{sec:3.4}
After generating parameters in the subtask processor, the selected tool and its parameters are passed to the executor. The executor operates independently of LLMs, focusing on the execution of specified tools.


\section{Experiments} \label{exp}
In this section, we outline the preparations of the entire experiment in Section \ref{exp1}, further details of experimental settings can be found in Appendix \ref{apd:first}. 
Then we empirically verify the efficiency and effectiveness of TUMS in Section \ref{exp2} with a series of ablation studies, focusing primarily on answering three questions, (1) Can the framework effectively tackle tasks of different difficulties with various tools? (2) Is parameter-level processing with multi-structure handlers significantly useful? (3) What role does the intent recognizer actually play in the framework?
We designed mainly three experiments as follows to answer the key questions mentioned above:
\begin{enumerate}[(1)]
    \item \textbf{TUMS vs. Baselines} Evaluate whether TUMS can effectively answer the questions, which the metric is the correct rate.
    \item \textbf{Multi-structure vs. One-Structure} TUMS incorporates multi-structure handlers. To demonstrate the advantages of these multi-structure handlers, we evaluate their performance in comparison to one-structure handlers.
    \item \textbf{Intent Recognizer vs. w/o Intent Recognizer} We also demonstrate the relationship between intent recognizer and the efficiency of TUMS.
\end{enumerate}

\subsection{Preparations} \label{exp1}
\subsubsection{Benchmark}
Our experiments are conducted on the ToolQA\cite{zhuang2024toolqa} benchmark, which comprises eight datasets spanning diverse domains and contains a total of 1,530 questions. These datasets serve as queryable external knowledge sources, encompassing text corpora, tabular databases, graphs, and mathematical problems.
For each dataset, ToolQA has designed questions at two difficulty levels. 
Both cannot be answered directly with LLMs' internal knowledge, but instead require LLMs to obtain information from the reference datasets via tool use. 

\subsubsection{Baselines}

We utilize several baseline methods, all based on the \texttt{Qwen1.5-72B}\cite{qwen1.5} model, to compare with the performance of our proposed approach:
(1) \textbf{Qwen}, where questions are fed directly into the Qwen model, and its output is taken as the final answer. (2) \textbf{CoT}\cite{wei2022cot}, which involves appending the phrase "Let's think step by step:" after each question to enhance the model's reasoning abilities. (3) \textbf{ReAct}\cite{yao2022react}, a method that boosts the problem-solving capabilities of LLMs by integrating reasoning with tool usage, prompting the models to produce alternating verbal reasoning traces and tool invocations (4) \textbf{Chameleon}\cite{lu2024chameleon}, an effective technique that employs LLMs as controllers to address subtasks with multiple tools.
ReAct and Chameleon methods can make use of the 13 tools provided by ToolQA\cite{zhuang2024toolqa}, which encompass text tools, database tools, math tools, graph tools, code tools, and system tools.

\subsection{Results and Analysis} \label{exp2}

\subsubsection{TUMS vs. Baselines} 
Table \ref{tab:easy_results} and Table \ref{tab:hard_results} show the discrepancy in performance among TUMS and baselines.
Our approach is distinguished by a gray background, with the bolded indicating the optimal performance for each dataset.

Since we do not provide the tools for Qwen and CoT(Qwen) methods, it is reasonable to assume the correct rates of the Qwen and CoT(Qwen) methods are both zero on the datasets except GSM8K, since GSM8K questions can be answered with self-reasoning without any external tools. 

\begin{table*}[h]
    \centering
    \caption{The correct rate on easy questions.}
    \centerline{
    \resizebox{1.0\linewidth}{!}{
    \begin{tabular}{@{}ccccccccccc@{}} 
        \toprule
        \textbf{Model} & \textbf{Method} & \multicolumn{8}{c}{\textbf{Dataset}} & \textbf{Average} \\ \cmidrule{3-10}
        & & Flight & Coffee & Yelp &Airbnb & DBLP & SciREX & Agenda & GSM8K &  \\
        \midrule
        \multirow{4}{*}{\begin{tabular}[c]{@{}cc@{}} \textbf{GPT-3.5}\\\textbf{Turbo} \footnotemark \end{tabular}} 
        & GPT & 2.0 & 0.0 & 15.0 & 0.0 & 0.0 & 2.0 & 0.0 & 26.0 & 5.63 \\
        & CoT & 1.0 & 1.0 & 9.0 & 0.0 & 0.0 & 0.0 & 0.0 & 30.0 & 5.13 \\
        & Chameleon & 30.0 & 9.0 & 8.0 & 4.0 & 3.0 & 0.0 & 4.0 & 27.0 & 10.63 \\
        & ReAct & 48.0 & 81.0 & 64.0 & 29.0 & 23.0 & 2.0 & 24.0 & 23.0 & 36.75 \\
        \midrule
        \multirow{5}{*}{\begin{tabular}[c]{@{}cc@{}} \textbf{Qwen1.5}\\\textbf{72B-Chat} \end{tabular}}
        & Qwen & 0.0 & 0.0 & 0.0 & 0.0 & 0.0 & 0.0 & 0.0 & 15.0 & 1.88 \\
        & CoT & 0.0 & 0.0 & 0.0 & 0.0 & 0.0 & 0.0 & 0.0 & \textbf{50.0} & 6.25 \\
        & Chameleon & 19.0 & 45.0 & 3.0 & 1.0 & 1.0 & 2.0 & 42.0 & 1.0 & 14.25 \\
        & ReAct & 45.0 & \textbf{88.0} & 67.0 & 60.0 & 24.0 & 0.0 & 40.0 & 49.0 & 46.63 \\
        & \cellcolor{gray!25}{TUMS (ours)}
        & \cellcolor{gray!25}\textbf{58.0}
        & \cellcolor{gray!25}\textbf{88.0}
        & \cellcolor{gray!25}\textbf{76.0}
        & \cellcolor{gray!25}\textbf{83.0}
        & \cellcolor{gray!25}\textbf{28.0}
        & \cellcolor{gray!25}\textbf{3.0}
        & \cellcolor{gray!25}\textbf{60.0}
        & \cellcolor{gray!25}\textbf{50.0}
        & \cellcolor{gray!25}\textbf{55.75} \\
        & & \cellcolor{green1!50}\scriptsize{$\Uparrow$28.9\%} 
        & - 
        & \cellcolor{green1!50}\scriptsize{$\Uparrow$13.4\%} 
        & \cellcolor{green1!50}\scriptsize{$\Uparrow$38.3\%} 
        & \cellcolor{green1!50}\scriptsize{$\Uparrow$16.7\%} 
        & \cellcolor{green1!50}\scriptsize{$\Uparrow$50.0\%} 
        & \cellcolor{green1!50}\scriptsize{$\Uparrow$42.9\%} 
        & - 
        & \cellcolor{green1!50}\scriptsize{$\Uparrow$19.6\%} \\
        \hline
        \bottomrule
    \end{tabular}\label{tab:easy_results}
    }}
\end{table*}

\begin{table*}[h]
    \centering
    \caption{The correct rate on hard questions.}
    \centerline{
    \resizebox{1.0\linewidth}{!}{
    \begin{tabular}{@{}cccccccccc@{}} 
        \toprule
        \textbf{Model} & \textbf{Method} & \multicolumn{7}{c}{\textbf{Dataset}} & \textbf{Average} \\ \cmidrule{3-9}
        & & Flight & Coffee & Yelp &Airbnb & DBLP & SciREX & Agenda &  \\
        \midrule
        \multirow{4}{*}{\begin{tabular}[c]{@{}cc@{}} \textbf{GPT-3.5}\\\textbf{Turbo} \footnote[7]{} \end{tabular}} 
        & GPT & 2.0 & 2.3 & 0.0 & 2.0 & 4.0 & 3.0 & 1.0 & 2.04 \\
        & CoT & 0.0 & 0.8 & 1.0 & 0.0 & 3.0 & 5.0 & 0.0 & 1.40 \\
        & Chameleon & 3.0 & 2.3 & 0.0 & 0.0 & 8.0 & 0.0 & 0.0 & 1.90 \\
        & ReAct & 5.0 & 17.7 & 8.0 & 7.0 & 5.0 & 8.0 & 7.0 & 8.24 \\
        \midrule
        \multirow{5}{*}{\begin{tabular}[c]{@{}cc@{}} \textbf{Qwen1.5}\\\textbf{72B-Chat} \end{tabular}}
        & Qwen & 0.0 & 0.0 & 0.0 & 0.0 & 0.0 & 0.0 & 0.0 & 0.0 \\
        & CoT & 0.0 & 0.0 & 0.0 & 0.0 & 0.0 & 0.0 & 0.0 & 0.0 \\
        & Chameleon & 0.0 & 2.3 & 0.0 & 0.0 & 1.0 & 10.0 & 0.0 & 1.90 \\
        & ReAct & \textbf{11.0} & 28.5 & 14.0 & \textbf{10.0} & 12.0 & 1.0 & 0.0 & 10.92 \\
        & \cellcolor{gray!25}{TUMS (ours)} 
        & \cellcolor{gray!25}6.0 
        & \cellcolor{gray!25}\textbf{36.2}
        & \cellcolor{gray!25}\textbf{36.0}
        & \cellcolor{gray!25}7.0
        & \cellcolor{gray!25}\textbf{16.0}
        & \cellcolor{gray!25}\textbf{14.0}
        & \cellcolor{gray!25}0.0
        & \cellcolor{gray!25}\textbf{16.45} \\
        & & \cellcolor{red1!50}\scriptsize{$\Downarrow$45.5\%} 
        & \cellcolor{green1!50}\scriptsize{$\Uparrow$27.0\%} 
        & \cellcolor{green1!50}\scriptsize{$\Uparrow$157.1\%} 
        & \cellcolor{red1!50}\scriptsize{$\Downarrow$30.0\%} 
        & \cellcolor{green1!50}\scriptsize{$\Uparrow$33.3\%} 
        & \cellcolor{green1!50}\scriptsize{$\Uparrow$40.0\%} 
        & - 
        & \cellcolor{green1!50}\scriptsize{$\Uparrow$50.6\%} \\
        \hline
        \bottomrule
    \end{tabular}\label{tab:hard_results}
    }}
\end{table*}

From the result results, we can find that the CoT(3-shot) method and TUMS achieve the best performance on GSM8K. 
For the two retrieval datasets, TUMS reached the peak performance, increased by approximately 42.9\% of existing methods on the Agenda-easy dataset, while all methods still have a poor performance on SciREX dataset, only Chameleon and TUMS achieve 10\% and 14\% separately on the SciREX-hard dataset. The explanation for this phenomenon is that the SciREX dataset collected from various full-length scientific papers is challenging, simultaneously the available retrieval tool only returns the first three most related contents. The latter is precisely the underlying cause of the failure of all methods on Agenda-hard.
For the DBLP dataset, TUMS gained an improvement of about 16.7\% on DBLP-easy and about 33.3\% on DBLP-hard.
It can be observed from the above results that TUMS can master the easy questions in four table datasets with marked advancement, while it shows an unstable performance in hard questions of table datasets. We show a detailed analysis of this phenomenon and propose a preference-based trick to mitigate this issue in Appendix \ref{apd:second}.

\footnotetext{The experiment results based on GPT-3.5-Turbo are from ToolQA\cite{zhuang2024toolqa}.}

\subsubsection{Multi vs. One}
The key contribution of our work is the parameter-level handlers with multiple structures, choosing according to diverse levels of difficulties of tool usage, specifically, the complexity of generating parameters of tools. 

\begin{figure}[h] \centering
\begin{subfigure}[Results on Easy Questions]{0.49\textwidth}
    \includegraphics[width=\textwidth]{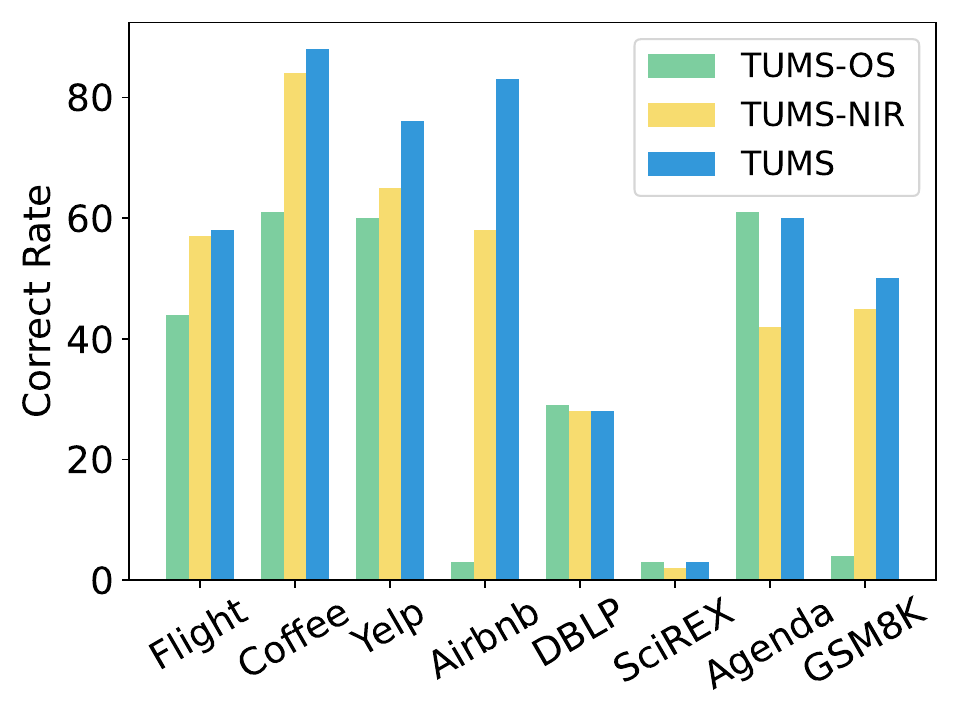}
\end{subfigure}
\begin{subfigure}[Results on Hard Questions]{0.49\textwidth}
    \includegraphics[width=\textwidth]{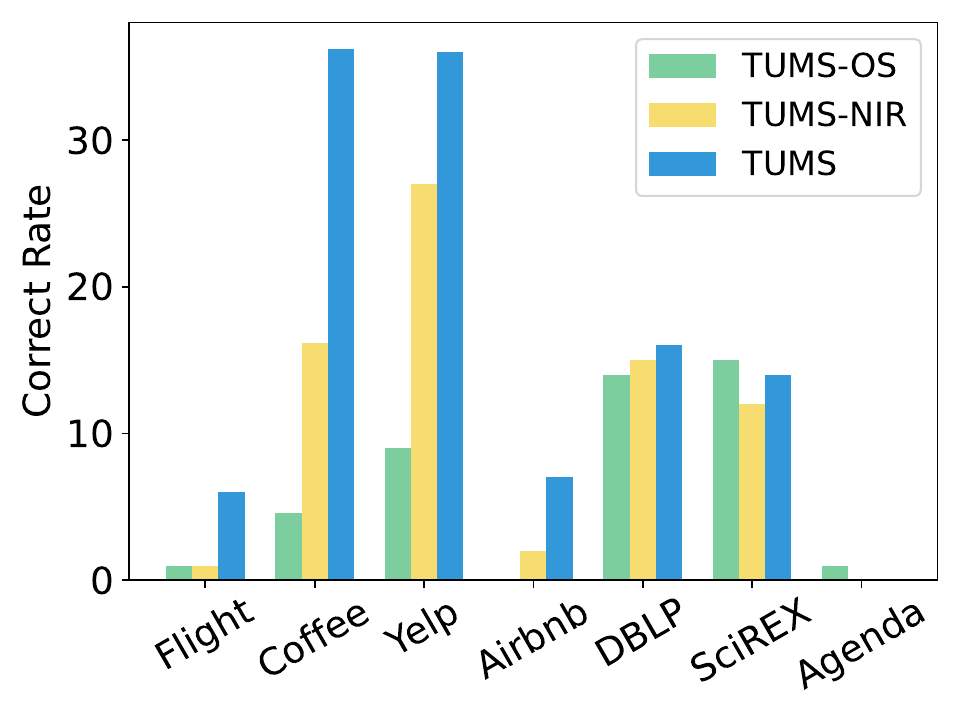}
\end{subfigure}
\caption{Comparison of TUMS Variants. TUMS is our proposed method; TUMS-NIR indicates TUMS without intent recognizer; TUMS-OS is the same as TUMS except it has only one-structure handlers.}
\label{fig:exp_variants}
\end{figure}

We experimentally demonstrated the significance of multi-structure in enhancing performance in Figure \ref{fig:exp_variants}. 
The one-structure TUMS, a variant of TUMS, is implemented using the same exemplars as TUMS differing in that it only employs a Direct Generation Structure. There is a noticeable overall downward across all datasets with about 19.9\% correct rate, exhibiting a poorer performance compared to ReAct method. TUMS, constructing multiple structures by considering the parameter-generation difficulties of various tools, can outperform all existing methods. We can conclude that the crucial addition to the framework is the fine-grained multi-structure handlers. 

\subsubsection{Recognizer vs. w/o Recognizer} \label{sec:exp3}
The likely biggest concern with TUMS is inefficient due to the "too-sophisticated" framework. In this part, we dispel the concern by visualizing the efficiency differences of ReAct, TUMS-NIR, and TUMS.

\begin{figure}[!h]
\centering
\begin{minipage}[b]{0.48\textwidth}
    We conducted statistics on the number of total responses from LLM, quantifying the cost of the corresponding method as a metric - \textit{total cost}. Dividing the total responses by the number of correct answers, we inferred the needed average number of responses on the correct as a metric - \textit{average cost}, representing the method's efficiency.
    As the illustration shown in Figure \ref{fig:exp_query}, TUMS without intent recognizer can still obtain decent average cost, which implies efficiency of task decomposer and multi-structure processor. By adding the recognizer, TUMS can significantly improve the efficiency with the lowest average cost despite having higher total query consumption which is mainly due to the inevitable queries from the recognizer. Generally speaking, the cost of adding recognizer is necessary since such markable performance improvement. 
\end{minipage}
\hfill
\begin{minipage}[b]{0.48\textwidth}
    \centering
    \includegraphics[width=\linewidth]{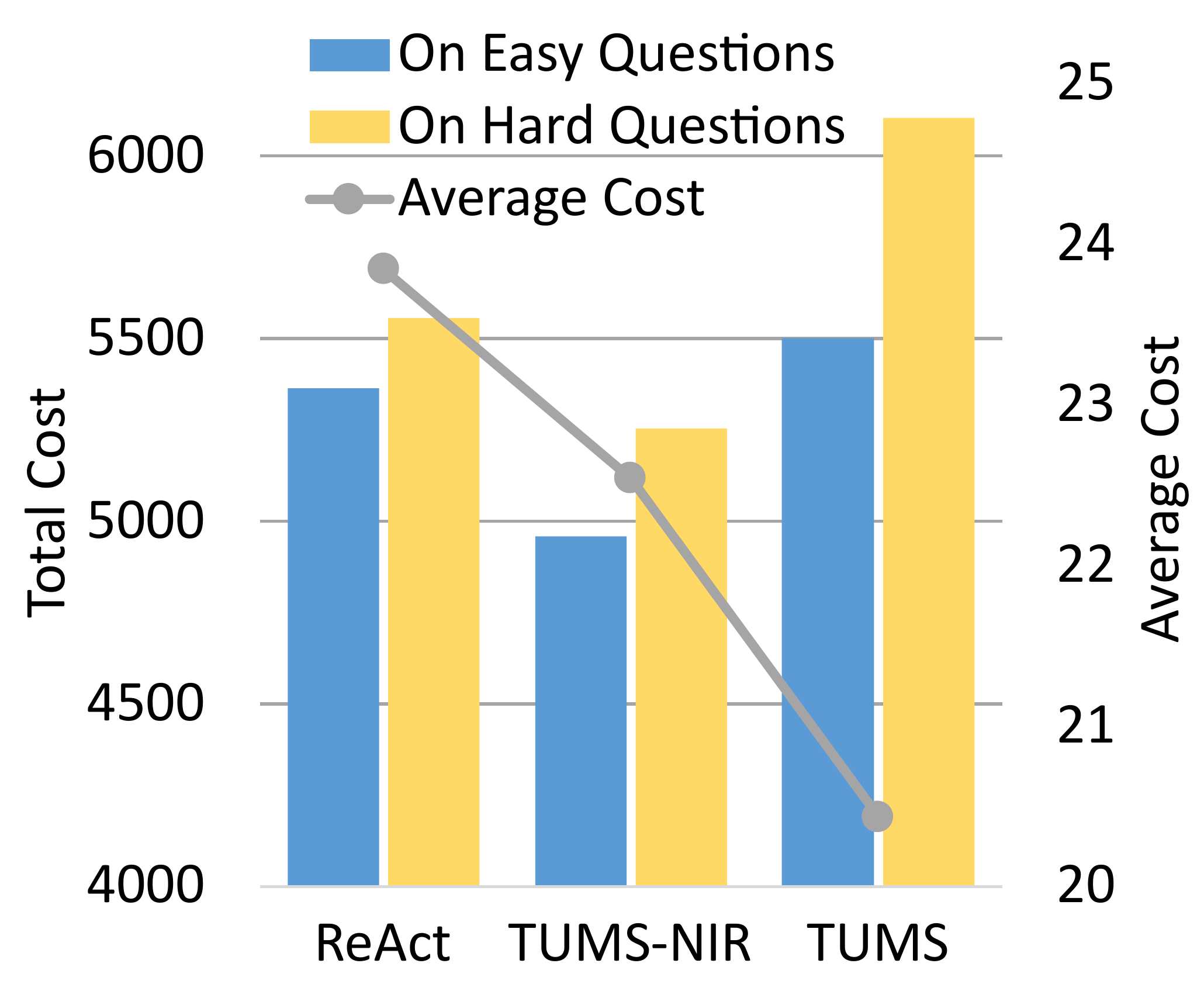}
    \caption{The analysis of efficiency. The total cost on the left y-axis(bar chart) and the average cost for one correct answer on the right y-axis(line chart), against different methods on the x-axis.}
    \label{fig:exp_query}
\end{minipage}
\end{figure}

\section{Conclusion}
In this paper, we have introduced TUMS, a multi-module collaboration framework that can enhance the tool-use capabilities of LLMs.
The proposed framework consists of four modules: the intent recognizer, the task decomposer, the subtask processor and the executor.
To overcome the confines of coarse-grained parameter generation, we propose multi-structure handlers in the subtask processor, which refines from tool-level processing to parameter-level processing.
We showed that the TUMS framework outperforms baselines in various tool-needed Question-Answering tasks and hope this work can provide a step toward the autonomous agent with powerful tool-use ability.

\begin{credits}
\subsubsection{\ackname} 
I would like to thank my supervisor, Bo Xu, for his guidance during this research. Your insightful feedback brought my work to a higher level. 
Thanks also to my mentor, Yanna Wang, for her guidance and valuable advice for this work. Your insight and thoughtful comments matter steered me through this research.


\subsubsection{\discintname}
The authors have no competing interests to declare that are relevant to the content of this article.

\end{credits}

\section{Appendix}
\subsection{Details of Experiments} \label{apd:first}
For all methods, the temperature of Qwen1.5-72B-Chat is set to 0, and the max-token is limited to 256 for each response. The random seed is set to 0. The other parameters for conducting experiments and ablations are the same as the settings in ToolQA\cite{zhuang2024toolqa}. We have open-sourced our code in GitHub.

\subsection{Preference-based Hint}\label{apd:second}
The outstanding performance of TUMS is attributed to the fine-grained parameter generation of multi-structure handlers in the subtask processor module, while the planning ability of tools primarily depends on the task decomposer. Therefore, facing the extremely challenging task, there is a bottleneck to generating an accurate plan of tools before manipulating them properly. This is a fairly feasible explanation for the mediocre performance of TUMS in some hard questions, shown in Table \ref{tab:hard_results}.

Since multiple tool-use paths may exist to complete a specific task, we can intuitively employ preference-based prior knowledge 
to guide the response of LLM. For example, for the question of \textbf{What is the total price at least if you want to stay at Bright, modern room with panoramic window in Maspeth for 8 nights?}, we can load the related database, filter, and then get the value of according price, at last calculate the answer with calculator tool. As an alternative option, using the SQLInterpreter tool with a moderately complex query statement can solve the problem. 
Our preference policy is that the shorter solving path is the better solving path, implemented by inducing LLM with a specific natural language hint, and the illustration is as follows.

\begin{figure}[!hp]
    \centering
    \input{iconip2024/prompts-tums}
    \input{iconip2024/prompts-hint}
    \caption{The prompts for all modules and two hints examples.}
    \label{fig:prompts_hints}
\end{figure}

%
%
\bibliographystyle{splncs04}
\bibliography{iconip2024}

\end{document}